\documentclass[runningheads]{llncs}
\usepackage{graphicx}
\usepackage[colorlinks=true, citecolor=black, linkcolor=black]{hyperref}

\newcommand{\refalgorithm}[1]{Algorithm~\ref{#1}}
\newcommand{\refequation}[1]{Equation~\ref{#1}}
\newcommand{\reffigure}[1]{\figurename~\ref{#1}}
\newcommand{\refsection}[1]{Section~\ref{#1}}

\newcommand{\reftable}[1]{Table~\ref{#1}}   

\usepackage{algorithm}
\usepackage{algpseudocode}
\usepackage{booktabs}
\usepackage{float}



\begin{document}
\title{Optimised Playout Implementations for the Ludii General Game System
}
%
%
\author{Dennis J. N. J. Soemers \and {\'E}ric Piette \and Matthew Stephenson \and Cameron Browne}
\authorrunning{D.J.N.J. Soemers et al.}
%
\institute{Department of Data Science and Knowledge Engineering, Maastricht University, Paul-Henri Spaaklaan 1, 6229 EN, Maastricht, the Netherlands
\email{\{dennis.soemers,eric.piette,matthew.stephenson,cameron.browne\}\\@maastrichtuniversity.nl}}
\maketitle              
\begin{abstract}
This paper describes three different optimised implementations of playouts, as commonly used by game-playing algorithms such as Monte-Carlo Tree Search. Each of the optimised implementations is applicable only to specific sets of games, based on their rules. The Ludii general game system can automatically infer, based on a game's description in its general game description language, whether any optimised implementations are applicable. An empirical evaluation demonstrates major speedups over a standard implementation, with a median result of running playouts 5.08 times as fast, over 145 different games in Ludii for which one of the optimised implementations is applicable.

\end{abstract}

\section{Introduction}

The playing strength of automated game-playing agents based on tree search algorithms, such as $\alpha\beta$-pruning \cite{Knuth_1975_AlphaBeta} and Monte-Carlo Tree Search (MCTS) \cite{Kocsis_2006_Bandit_Abbrev,Coulom_2007_MCTS,Browne_2012_MCTS_Abbrev}, typically correlates strongly with the efficiency of basic operations such as computing a list of legal moves, applying a move to a state, copying a game state, or evaluating whether or not a state is terminal. When such operations can be implemented to run more efficiently, they allow for deeper tree searches, which usually leads to stronger agents. For this reason, a significant amount of research has gone towards techniques such as bitboard methods \cite{Browne_2014_Bitboard}, PropNet optimisations \cite{Sironi_2017_Optimizing} for general game playing, hardware accelerators \cite{Campbell_2002_DeepBlue,Siwek_2018_FPGA_Abbrev}, optimising compilers for general game description languages \cite{Kowalski_2020_Efficient_Abbrev}, etc.

MCTS is one of the most commonly used tree search algorithms for general game playing \cite{Finnsson_2010_Learning_Abbrev,Swiechowski_2015_RecentAdvancesGGP}. Typically, a significant portion of the time spent by this algorithm is in running \textit{playouts}; these may intuitively be understood as the algorithm following a ``narrow'' and ``deep'' trajectory of several---often many---consecutive states and actions. In their most basic form, playouts are run by selecting legal actions uniformly at random, and continuing them until a terminal game state is reached, but it is also possible to truncate playouts early and to select actions during playouts according to non-uniform distributions.

After running a playout, it is typically not necessary to retain the intermediate states generated between the start and end of a playout, the lists of legal moves, etc.; only the final outcome of a playout is generally of interest. This is in contrast to minimax-based algorithms such as $\alpha\beta$-pruning \cite{Knuth_1975_AlphaBeta}, or even the time spent by MCTS in its tree building and traversal (outside of playouts), where intermediate states and exact lists of legal moves are required for a correct tree to be built. Straightforward playout implementations compute exact lists of legal moves in every state anyway, such that actions may be sampled from them afterwards, but these insights may be used to develop more efficient playout implementations.

In this paper, we propose several different optimised playout implementations for the Ludii general game system \cite{Browne_2020_Practical_Abbrev,Piette_2020_Ludii}, which allow for playouts to be run significantly more quickly than with naive implementations. Each of them is only applicable to a restricted set of games, but the system can automatically determine for any given game whether or not any specific playout implementation is applicable. Furthermore, each of the proposed implementations is applicable to a substantial number of games in Ludii (i.e., not specific to just a single or a handful of games). Only our approach for automatically determining the applicability of playout implementations is specific to Ludii---in particular, to its game description format. The basic ideas behind the optimised playout implementations are not specific to Ludii, and may be relevant for other general game systems as well as single-game engines.

\section{Background}

Ludii is a general game system that can run any game described in its \textit{ludemic} game description format \cite{Browne_2020_Practical_Abbrev,Piette_2020_Ludii}. A large library of \textit{ludemes}, which may intuitively be understood as keywords that make up the game description language, is automatically inferred from Ludii's codebase using a class grammar approach \cite{Browne_2016_Class_Abbrev}. An example game description for the game of Tic-Tac-Toe in Ludii's game description language is provided by \reffigure{Fig:TicTacToeDescription}.
\begin{figure}
\centering
{\tt \footnotesize
\begin{verbatim}
    (game "Tic-Tac-Toe"  
        (players 2)  
        (equipment { 
            (board (square 3)) 
            (piece "Disc" P1) 
            (piece "Cross" P2) 
        })  
        (rules 
            (play (move Add (to (sites Empty))))
            (end (if (is Line 3) (result Mover Win)))
        )
    )
\end{verbatim}
}
\caption{Game description for Tic-Tac-Toe in Ludii's game description language.}
\label{Fig:TicTacToeDescription}
\end{figure}

Any game described in this language can be compiled by Ludii, resulting in a forward model with functions for computing lists of legal moves, applying moves to game states, copying game states, etc. Given these functions, a straightforward playout implementation can be written as in \refalgorithm{Alg:NaivePlayout}.
\begin{algorithm}
\begin{algorithmic}[1]
\Require Game state $s$ to start playout from.
\While{playout should be continued} \Comment{Not terminal and not truncated}
    \State \texttt{legal\_moves} $\gets$ \Call{ComputeLegalMoves}{$s$}
    \State Sample move $m$ from \texttt{legal\_moves} \Comment{Often uniformly at random}
    \State Apply move $m$ to state $s$
\EndWhile
\State \Return game state $s$ at end of playout.
\end{algorithmic}
\caption{Standard playout implementation.} \label{Alg:NaivePlayout}
\end{algorithm}

\section{Related Work}

For several connection games \cite{Raiko_2008_UCT_Abbrev} (and possibly other types of games), it can be proven that a game always ends in a win for exactly one player (no ties), and that the outcome does not change if play ``continues'' after reaching a terminal game state until the game board is full. For such games, playouts can be optimised by simply continuing them until the board is full, and only evaluating the outcome once at the end \cite{Raiko_2008_UCT_Abbrev}. This is efficient because evaluating the win condition, which is often the most expensive computation of these games, only needs to be done once, at the end of every playout. This is in contrast to standard playout implementations as in \refalgorithm{Alg:NaivePlayout}, where the win condition would be evaluated after every move.

In a general game system such as Ludii, we do not have a straightforward way to automatically prove or disprove for any arbitrary game description that the properties required for the optimisation described above hold.
However, the techniques we propose in the following sections are similar in the sense that they are tailored specifically towards optimising playouts, as opposed to more generally optimising functions that are also used outside of playouts.

\section{Add-to-Empty Playouts} \label{Sec:AddToEmptyPlayouts}

The first collection of games for which we propose an optimised playout implementation is the set of games where players' moves consist of placing pieces of their colour on empty positions on a game board, and pieces can never be moved or removed anymore after being placed. We refer to these as ``add-to-empty'' games. This includes many well-known games such as \textit{Gomoku}, \textit{Havannah}, \textit{Hex}, \textit{Tic-Tac-Toe}, \textit{Yavalath}, etc. These are often connection or line-completion games.

More formally, in Ludii, these games are recognised as those games where the playing rules are defined as \texttt{(play (move Add (to (sites Empty))))}. This is a strong restriction because only a single specific set of playing rules is permitted, but in practice we find this particular ruleset to be relatively commonly used among several popular games. For this specific set of rules, we are guaranteed that the list of legal moves in the initial game state is simply represented by all positions that are empty at the start of the game (generally the entire board), and that this list of legal moves monotonically decreases by exactly one after every move. This allows for an optimised implementation, where the list of legal moves is pre-allocated once at the start of a playout, and legal moves do not need to be re-computed at any later stage in the same playout.

The only exception that we implement additional support for is the \textit{swap rule} (or pie rule). This is a common rule used in many of the games we aim to cover with this playout implementation, such as Hex and Havannah, which states that in the first turn of the second player, that player may opt to swap colours with their opponent, rather than making a move. 
This rule is intended to eliminate a first-mover advantage that the first player otherwise often has in these games. 
The presence of this rule technically means that the list of moves does not monotonically decrease by one in the very first turn transition, but it is straightforward to implement support for this one special case in the optimised playout implementation.

Note that, in these games, the idea of pre-computing a list of legal moves only once at the start, and monotonically removing moves as they are played afterwards, does not necessarily have to be restricted to just playouts. If such a list of moves were stored in memory in the game state representation, and updated as moves were applied, the optimisation could also be used outside of playouts (e.g., when building search trees). In the Regular Boardgames system \cite{Kowalski_2019_Regular_Abbrev}, such an idea has been implemented more generally as a step of an optimising compiler \cite{Kowalski_2020_Efficient_Abbrev}. However, we remark that this does increase the memory footprint of the game state representation, and it can slow down operations such as the copying of game states, which is often required in aspects of game tree searches outside of playouts.
\footnote{Ludii often requires game states to be copied during tree searches because Ludii does not support ``undoing'' moves, though this may be added in the future.} 
By restricting the use of this idea to just playouts, where generating intermediate copies of game states is not required, we are guaranteed that it cannot inadvertently cause a slowdown.

\section{Filter Playouts} \label{Sec:FilterPlayouts}

The second collection of games for which we provide an optimised playout implementation is the set of games where there is a basic set of arbitrary rules that defines an initial list of legal moves for any game state $s$, but some of these moves $m$ are afterwards filtered out if a certain postcondition fails for whichever successor state $s'$ is reached if $m$ were to be applied to $s$. A well-known example of such a game is \textit{Chess}, where at first the moves are described according to the different move rules of different pieces, but any move $m$ that would lead to a successor state $s'$ where the mover's king would (still) be under threat is filtered out. In chess-specific engines, such conditions may be relatively cheap to compute without actually generating all the hypothetical successor states $s'$. However, in the Ludii general game system, these conditions are expensive to compute because all the potential successor states $s'$ are fully generated (which in turn first requires many copies of $s$ to be generated) to evaluate the postconditions.

More formally, we provide support for any game in Ludii where the playing rules are described in any one of the following formats, where isolated capital letters \texttt{A}, \texttt{B}, etc. can be filled by any arbitrary rules as permitted by the game description language:
\begin{enumerate}
    \item \texttt{(play (do A ifAfterwards:(B)))}
    \item \texttt{(play (if A B (do C ifAfterwards:(D))))}
    \item \texttt{(play (or (do A ifAfterwards:(B)) (move Pass)))}
\end{enumerate}
The first case is the most basic case, where \texttt{A} defines the rules used to generate the unfiltered list of moves, and \texttt{B} defines the postcondition that must hold in the successor state for any move generated by \texttt{A} not to be filtered out. The second case generates moves according to \texttt{B} if condition \texttt{A} holds, and otherwise drops into a similar construction as in the first case. This construction is frequently used in games such as \textit{Chess} and \textit{Shogi}, where promotion moves are generated if the player to make a move is the same player as the last mover, and regular moves with postconditions are generated otherwise. The third case is similar to the first case, except it also always generates an unconditional pass move as a legal move. This is used for games such as \textit{Go}, where placing stones is conditional on liberty postconditions, but passing is always permitted. Other (more complex) cases than these three may occur and could be supported, but adding such support would require a small amount of additional engineering effort on a case-by-case basis. In practice we found these three cases to provide sufficient coverage for a substantial number of games, including several popular ones such as Chess, Go, and Shogi.

When constructing game trees, we cannot avoid computing the expensive postconditions, because the exact lists of legal moves must be fully generated to construct a correct game tree. However, in playouts, we only require the ability to sample legal moves according to some desired distribution over the legal moves, but do not necessarily need to know which other (unsampled) moves were actually legal according to the postconditions. Hence, we propose a playout implementation where moves are generated without checking postconditions. A rejection sampling approach is used where postconditions are evaluated only after a move has been selected (uniformly at random, in the simplest case), and the process is repeated if it turns out that the sampled move should have been filtered out. This allows us to avoid evaluating potentially expensive postconditions for moves that are not sampled. Pseudocode for this approach is provided by \refalgorithm{Alg:FilterPlayout}. \refsection{Sec:NonuniformDistributions} discusses how this approach can be combined with more sophisticated playouts with non-uniform distributions over moves.
\begin{algorithm}
\begin{algorithmic}[1]
\Require Game state $s$ to start playout from.
\While{playout should be continued} \Comment{Not terminal and not truncated}
    \State \texttt{moves} $\gets$ \Call{ComputeMaybeLegalMoves}{$s$} \Comment{Ignore postconditions}
    \State $m \gets $ \textsc{Null}
    \While{$m = $ \textsc{Null}}
        \State $m \gets$ sample move $m$ from \texttt{moves}
        \If{$m$ fails postcondition}
            \State $m \gets $ \textsc{Null}
            \State Remove $m$ from \texttt{moves}
        \EndIf
    \EndWhile
    \State Apply move $m$ to state $s$
\EndWhile
\State \Return game state $s$ at end of playout.
\end{algorithmic}
\caption{Optimised filter playout.} \label{Alg:FilterPlayout}
\end{algorithm}

\section{No-Repetition Playouts} \label{Sec:NoRepetitionPlayouts}

The final playout implementation we propose is a variant of the filter playouts described in the previous section. Outside of the general playing rules, Ludii's game description language also allows for a more general \texttt{(noRepeat)} ``meta-rule'' to be applied to a complete game. 
When this rule is used, any move that leads to a game state that has already been encountered before is illegal.
This can be viewed as an additional postcondition, which again requires a game state copy and a move application to evaluate, as described in \refsection{Sec:FilterPlayouts}. A similar rejection sampling approach can also be used again to avoid these computations for many legal moves in playouts. The main difference between the no-repetition playout and the filter playout is simply in how its applicability can be determined from a game's game description file. In games where filter playouts are also valid, any repetition restrictions are evaluated at the same time as the optimised postconditions.

\section{Non-uniform Move Distributions} \label{Sec:NonuniformDistributions}

Selecting moves uniformly at random is a common and straightforward strategy, but it is often beneficial to use ``smarter'' playouts based on domain knowledge, offline learning, or online learning, which means that moves are sampled according to non-uniform distributions over the legal moves. The add-to-empty playouts described in \refsection{Sec:AddToEmptyPlayouts} still generate the precise lists of legal moves, which means that they support the use of such non-uniform distributions. However, the filter playouts and no-repetition playouts described in Sections \ref{Sec:FilterPlayouts} and \ref{Sec:NoRepetitionPlayouts} require careful attention. These playout implementations may include illegal moves in their lists of moves, which are only discovered to be illegal and rejected after sampling them, but their presence in the initial list of moves may affect the probabilities computed for other (legal) moves. This may lead to an unintended change in the distribution over moves.

One common approach for move selection in playouts is to assign scores to moves, which are not translated into probabilities, but instead used to inform move selection through other means, such as $\epsilon$-greedy policies. An $\epsilon$-greedy strategy simply selects moves uniformly at random with probability $0 \leq \epsilon \leq 1$, or greedily with respect to the move scores with probability $1 - \epsilon$. Move scores can, for example, be obtained using approaches such as MAST, FAST \cite{Finnsson_2010_Learning_Abbrev}, NST \cite{Tak_2012_NGrams_Abbrev}, or PPA \cite{Cazenave_2015_PPA_Abbrev}. Techniques with only two or three discrete levels of prioritisation for moves, such as the Last-Good-Reply policy \cite{Baier_2010_Power_Abbrev} or decisive and anti-decisive moves \cite{Teytaud_2010_Decisive}, may be viewed as a special case with discrete move scores. Whenever such an $\epsilon$-greedy policy is used (including the special case of greedy policies with $\epsilon = 0$), our proposed playout implementations---with their rejection sampling schemes for handling illegal moves---will automatically play according to the correct (non-uniform) distributions, with no further changes required.

Another common approach is to compute a discrete probability distribution over all moves, and sample moves according to those probabilities. This is sometimes done by transforming move scores, such as those described above, into probabilities using a Boltzmann distribution. Given a set of legal moves $\mathcal{M}$, and a temperature hyperparameter $\tau$, the probability $p(m, \mathcal{M})$ with which a move $m \in \mathcal{M}$ with a score $Q(m)$ should be selected is then given by \refequation{Eq:Boltzmann}:
\begin{equation} \label{Eq:Boltzmann}
    p(m, \mathcal{M}) = \frac{\exp(Q(m) / \tau)}{\sum_{m' \in \mathcal{M}} \exp(Q(m') / \tau)}
\end{equation}
When offline training is used to train policies, for instance based on deep neural networks \cite{Silver_2018_AlphaZero} or simpler function approximators and state-action features \cite{Soemers_2019_Biasing_Abbrev}, it is also customary to use such a distribution with $\tau = 1$ (leading to a softmax distribution) and the $Q(m)$ values referred to as logits.

Let $\mathcal{M}$ denote a set of legal moves, and let $\mathcal{I}$ denote a set of moves as generated during a filter or no-repetition playout (which may include some illegal moves), such that $\mathcal{M} \subseteq \mathcal{I}$. Let $m_1$ and $m_2$ denote two arbitrary legal moves. The ratio $\frac{p(m_1, \mathcal{I})}{p(m_2, \mathcal{I})}$ between their probabilities, in the possible presence of illegal moves, is given by \refequation{Eq:RatioLegalMoves}:
\begin{equation} \label{Eq:RatioLegalMoves}
    \frac{p(m_1, \mathcal{I})}{p(m_2, \mathcal{I})} = \frac{\exp(Q(m_1) / \tau)}{\sum_{m' \in \mathcal{I}} \exp(Q(m') / \tau)} \times \frac{\sum_{m' \in \mathcal{I}} \exp(Q(m') / \tau)}{\exp(Q(m_2) / \tau)} = \frac{\exp(Q(m_1) / \tau)}{\exp(Q(m_2) / \tau)}
\end{equation}
Note that this ratio is equal to the ratio we would have had with $\mathcal{M}$ instead of $\mathcal{I}$, i.e. if there were no possible presence of illegal moves.

Let $m \in \mathcal{I}$ denote a move that has been sampled in a playout, and is rejected due to it turning out to be illegal, i.e. $m \notin \mathcal{M}$. For any other move $m' \neq m$, the probability value $p(m', \mathcal{I} \setminus \{m\})$ can be incrementally updated as $p(m', \mathcal{I} \setminus \{m\}) = p(m', \mathcal{I}) \times \frac{1}{1 - p(m, \mathcal{I})}$ when $m$ is rejected. This re-normalises the distribution into a proper probability distribution again after the rejection of the illegal move, without changing the ratio of probabilities between any pair of remaining moves, and without requiring the full distribution to be re-computed from scratch.

\section{Empirical Evaluation}

We evaluate the performance of the proposed playout implementations by measuring the average number of complete random playouts---from initial game state until terminal game state---that can be run per second, using both standard implementations (\refalgorithm{Alg:NaivePlayout}) and the optimised implementations. Every process is run on a single CPU core @2.2 GHz, using 60 seconds of warming up time for the Java Virtual Machine (JVM), followed by 600 seconds over which the number of playouts run per second is measured. We allocate 5120MB of memory per process, of which 4096MB is made available to the JVM. 

The version of Ludii used for this evaluation\footnote{Revision \texttt{7903697} of \url{https://github.com/Ludeme/Ludii}.} has 929 different games, with 1053 rulesets (some games can be played using several different variants of rules). Of these, 145 rulesets (from 141 games) are automatically detected to be compatible with one of the three proposed playout implementations. For each of them, we evaluate the speedup
as the number of playouts per second when using the optimised playout, divided by the number of playouts per second when using a standard playout implementation. For example, a speedup of $2.0$ means that the optimised implementation allows for playouts to be run twice as fast.

\begin{table}[t!]
\caption{Aggregate measures of the speedups obtained by different playout implementations in their applicable games.}
\begin{center}
\begin{tabular}{@{}l@{\hskip 12pt}c@{\hskip 12pt}r@{\hskip 12pt}r@{\hskip 12pt}r@{\hskip 12pt}r@{}}
\toprule
& & \multicolumn{4}{c}{Speedup} \\
\cmidrule(lr){3-6}
Playout Implementation & Num. Games & Min & Median & Mean & Max \\
\midrule
Add-To-Empty & 35 & 1.00 & 1.90 & 3.64 & 20.25 \\
Filter & 105 & 1.18 & 5.49 & 6.88 & 34.31 \\
No-Repetition & 5 & 1.65 & 6.35 & 9.08 & 19.26 \\
\midrule
All & 145 & 1.00 & 5.08 & 6.17 & 34.31 \\
\bottomrule
\end{tabular}
\label{Table:Results}
\end{center}
\end{table}
\begin{figure}[t!]
\centering
\includegraphics[width=\textwidth]{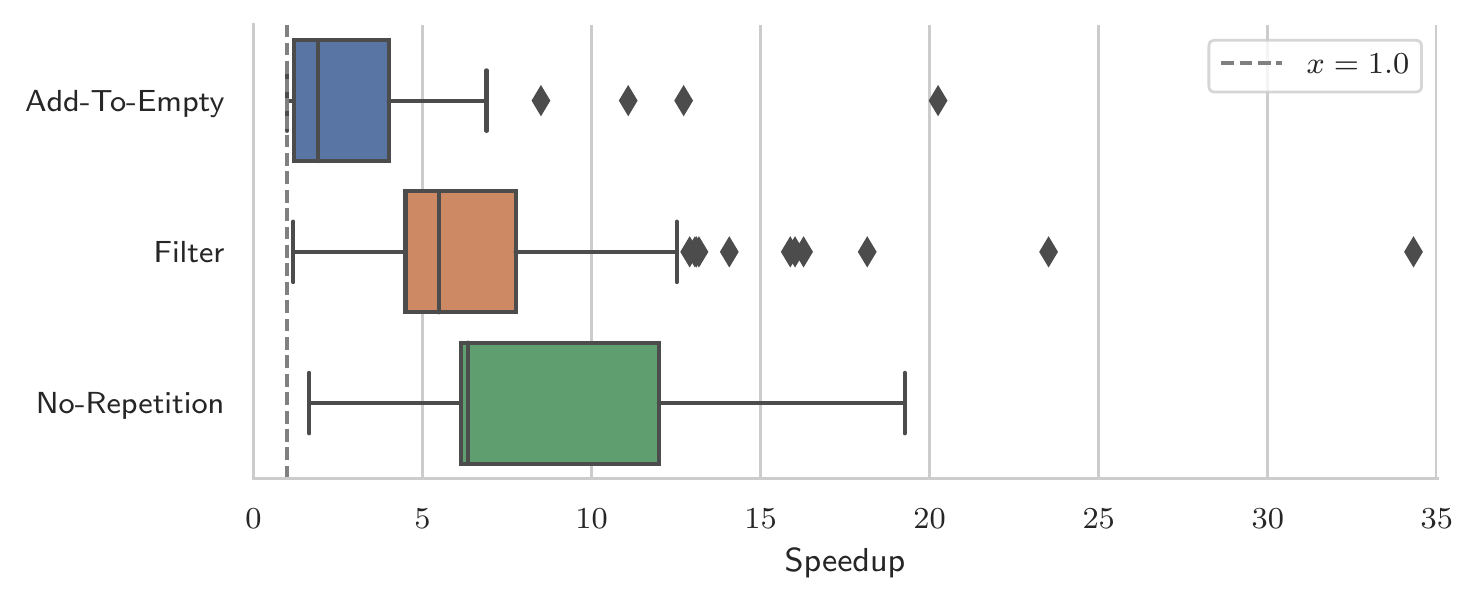}
\caption{Boxplots summarising speedups obtained from using optimised playout implementations rather than the standard one. Every data point is a different game (or ruleset). Points to the left of the $x = 1.0$ line are slowdowns.}
\label{Fig:Boxplots}
\end{figure}

\reffigure{Fig:Boxplots} summarises, for each of the three playout implementations, the different speedups obtained by using the optimised playout implementations in applicable games. \reftable{Table:Results} provides additional details on these results. Each of the three implementations provides noticeable speedups in the majority of games, with median speedups ranging from $1.90$ (almost twice as fast) for Add-To-Empty, to $6.35$ (more than six times faster) for No-Repetition. 
The largest speedup ($34.31$) is obtained by the Filter playout in the game of \textit{Go}. 

Only the Add-To-Empty playout has two games (out of 35) for which the speedup is lower than $1.0$, i.e. a slowdown; $0.9999896$ for \textit{Icosian}, and $0.997$ for \textit{Gyre}. In Icosian, the Add-To-Empty playout is only valid for the first phase of the game, which only lasts for a single move; after this phase, it is necessary to switch back to the standard playout implementation, and the overhead of this switch may cause the slowdown. In Gyre, close to $100\%$ of the time is spent computing the game's win condition, which is not affected by Add-To-Empty.

\begin{figure}[t!]
\centering
\includegraphics[width=.9\textwidth]{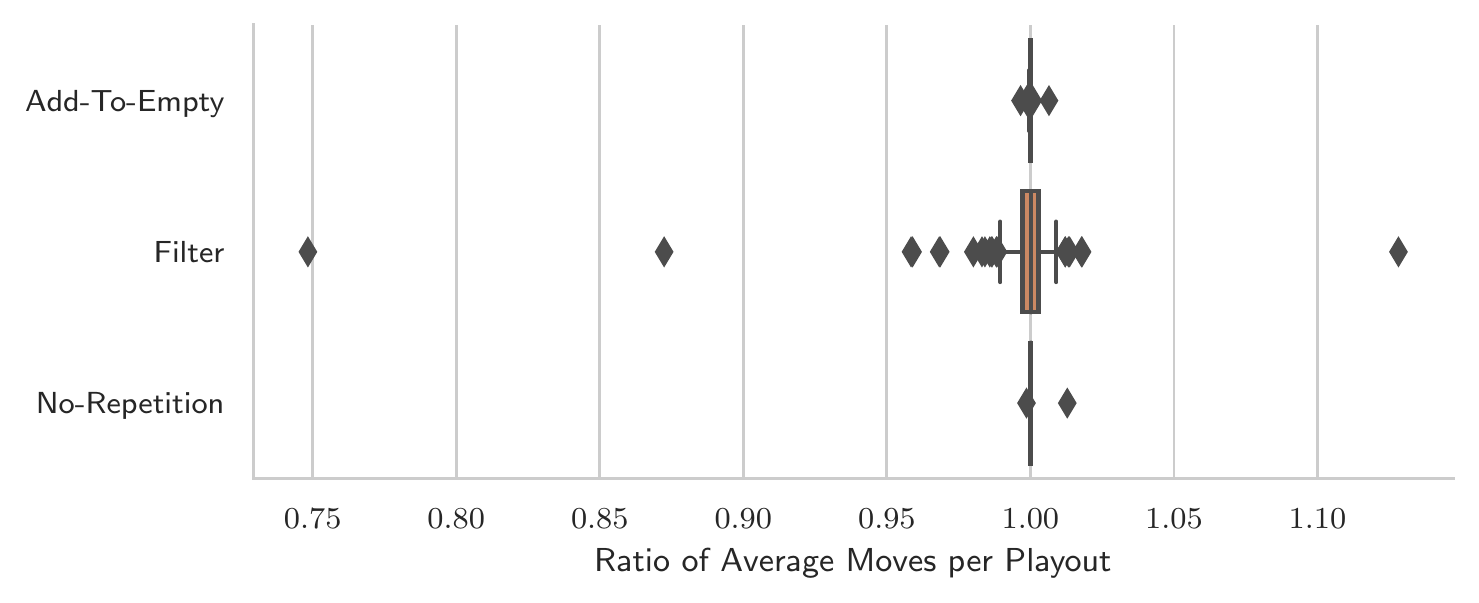}
\caption{For each of the optimised playout implementation, a boxplot summarising, for each game, the ratio between the recorded average numbers of moves per random playout with and without using the optimised implementation. Ratios less than $1.0$ mean that random playouts were shorter on average when using optimised implementations, and ratios greater than $1.0$ mean that random playouts were longer on average when using optimised implementations.}
\label{Fig:BoxplotsRelMoveChanges}
\end{figure}

In theory, the optimised playout implementations should not affect the probabilities with which moves are selected, and therefore random playouts should---on average---take equally long (measured in number of moves per playout) regardless of implementation. To verify that this is the case (i.e., there are no implementation errors), we compute a ratio for every game by dividing the average playout length recorded when using optimised implementations, by the corresponding number recorded when using the standard (unoptimised) implementation. The boxplots in \reffigure{Fig:BoxplotsRelMoveChanges} confirm that almost all these ratios are very close to $1.0$. 

The three biggest outliers are \textit{Hexshogi}, \textit{Unashogi}, and \textit{Yonin Shogi}, with ratios of $0.75$, $0.87$, and $1.13$, respectively. All three of these games are relatively slow games, which means that even in our $600$-second timing runs we obtain relatively low total numbers of playouts, with a significant variance in the number of moves per playout. Therefore, the observation of these outliers can be explained by a combination of relatively low sample sizes ($459$, $215$, and $322$ total playout counts over $600$ seconds for the three respective games when using optimised playout implementations) and high variance, rather than implementation errors. For all three of these outliers, the speedups recorded for the Filter playout are also more substantial than can be explained solely by the differences in average playout lengths; we record speedups of $7.52$, $5.67$, and $4.50$.

\section{Conclusion}

In this paper, we have proposed three optimised implementations for running playouts, as often used by algorithms such as MCTS. Each of the implementations is applicable to a specific set of games, depending on the rules used by a game. The Ludii general game system can automatically infer, based on game descriptions in its game description language, which---if any---of these implementations are applicable, and use them for running playouts when applicable. An empirical evaluation across 145 games demonstrated significant speedups, with a median result of running playouts $5.08$ times faster, a mean speedup of $6.17$ times, and a maximum speedup of $34.31$ times in the game of Go.

\section*{Acknowledgements}

This research is funded by the European Research Council as part of the Digital Ludeme Project (ERC Consolidator Grant \#771292) led by Cameron Browne at Maastricht University's Department of Data Science and Knowledge Engineering. We thank the anonymous reviewers for their feedback.

\bibliographystyle{splncs04}
\bibliography{dlp-biblio-1}

\begin{thebibliography}{10}
\providecommand{\url}[1]{\texttt{#1}}
\providecommand{\urlprefix}{URL }
\providecommand{\doi}[1]{https://doi.org/#1}

\bibitem{Baier_2010_Power_Abbrev}
Baier, H., Drake, P.D.: The power of forgetting: Improving the last-good-reply
  policy in monte carlo go. IEEE Trans. Comput. Intell. AI Games
  \textbf{2}(4),  303--309 (2010)

\bibitem{Browne_2014_Bitboard}
Browne, C.: Bitboard methods for games. ICGA Journal  \textbf{37}(2),  67--84
  (2014)

\bibitem{Browne_2012_MCTS_Abbrev}
Browne, C., Powley, E., Whitehouse, D., Lucas, S., Cowling, P.I., Rohlfshagen,
  P., Tavener, S., Perez, D., Samothrakis, S., Colton, S.: A {S}urvey of
  {M}onte {C}arlo {T}ree {S}earch {M}ethods. IEEE Trans. Comput. Intell. AI
  Games  \textbf{4}(1),  1--49 (2012)

\bibitem{Browne_2020_Practical_Abbrev}
Browne, C., Stephenson, M., Piette, {\'E}., Soemers, D.J.N.J.: A practical
  introduction to the ludii general game system. In: Cazenave, T., van~den
  Herik, J., Saffidine, A., Wu, I.C. (eds.) Adv. in Computer Games. ACG 2019.
  LNCS, vol. 12516. Springer, Cham (2020)

\bibitem{Browne_2016_Class_Abbrev}
Browne, C.B.: A class grammar for general games. In: Adv. in Computer Games.
  LNCS, vol. 10068, pp. 167--182. Leiden (2016)

\bibitem{Campbell_2002_DeepBlue}
Campbell, M., {Joseph Hoane Jr.}, A., Hsu, F.: Deep blue. Artificial
  Intelligence  \textbf{134}(1--2),  57--83 (2002)

\bibitem{Cazenave_2015_PPA_Abbrev}
Cazenave, T.: Playout policy adaptation for games. In: Plaat, A., van~den
  Herik, J., Kosters, W. (eds.) Adv. in Computer Games (ACG 2015). pp. 20--28.
  LNCS, Springer International Publishing (2015)

\bibitem{Coulom_2007_MCTS}
Coulom, R.: Efficient selectivity and backup operators in {M}onte-{C}arlo tree
  search. In: van~den Herik, H.J., Ciancarini, P., Donkers, H.H.L.M. (eds.)
  Computers and Games. LNCS, vol.~4630, pp. 72--83. Springer Berlin Heidelberg
  (2007)

\bibitem{Finnsson_2010_Learning_Abbrev}
Finnsson, H., Bj{\"o}rnsson, Y.: Learning simulation control in general
  game-playing agents. In: Proc. 24th AAAI Conf. Artificial Intell. pp.
  954--959. AAAI Press (2010)

\bibitem{Knuth_1975_AlphaBeta}
Knuth, D.E., Moore, R.W.: An analysis of alpha-beta pruning. Artificial
  Intelligence  \textbf{6}(4),  293--326 (1975)

\bibitem{Kocsis_2006_Bandit_Abbrev}
Kocsis, L., Szepesv{\'a}ri, C.: Bandit based {M}onte-{C}arlo planning. In:
  F{\"u}rnkranz, J., Scheffer, T., Spiliopoulou, M. (eds.) Mach. Learn.: ECML
  2006, LNCS, vol.~4212, pp. 282--293. Springer, Berlin, Heidelberg (2006)

\bibitem{Kowalski_2020_Efficient_Abbrev}
Kowalksi, J., Miernik, R., Mika, M., Pawlik, W., Sutowicz, J., Szyku{\l}a, M.,
  Tkaczyk, A.: Efficient reasoning in regular boardgames. In: Proc. 2020 IEEE
  Conf. Games. pp. 455--462. IEEE (2020)

\bibitem{Kowalski_2019_Regular_Abbrev}
Kowalski, J., Maksymilian, M., Sutowicz, J., Szyku{\l}a, M.: Regular
  boardgames. In: Proc. 33rd AAAI Conf. Artificial Intell. pp. 1699--1706. AAAI
  Press (2019)

\bibitem{Piette_2020_Ludii}
Piette, {\'E}., Soemers, D.J.N.J., Stephenson, M., Sironi, C.F., Winands,
  M.H.M., Browne, C.: Ludii -- the ludemic general game system. In: Giacomo,
  G.D., Catala, A., Dilkina, B., Milano, M., Barro, S., Bugarín, A., Lang, J.
  (eds.) Proceedings of the 24th European Conference on Artificial Intelligence
  (ECAI 2020). Frontiers in Artificial Intelligence and Applications, vol.~325,
  pp. 411--418. IOS Press (2020)

\bibitem{Raiko_2008_UCT_Abbrev}
Raiko, T., Peltonen, J.: Application of {UCT} search to the connection games of
  {H}ex, {Y}, *{S}tar, and {R}enkula! In: Proc. Finnish Artificial Intell.
  Conf. pp. 89--93 (2008)

\bibitem{Silver_2018_AlphaZero}
Silver, D., Hubert, T., Schrittwieser, J., Antonoglou, I., Lai, M., Guez, A.,
  Lanctot, M., Sifre, L., Kumaran, D., Graepel, T., Lillicrap, T., Simonyan,
  K., Hassabis, D.: A general reinforcement learning algorithm that masters
  chess, shogi, and {G}o through self-play. Science  \textbf{362}(6419),
  1140--1144 (2018)

\bibitem{Sironi_2017_Optimizing}
Sironi, C.F., Winands, M.H.M.: Optimizing propositional networks. In: Cazenave,
  T., Winands, M., Edelkamp, S., Schiffel, S., Thielscher, M., Togelius, J.
  (eds.) Computer Games. CGW 2016, GIGA 2016. Communications in Computer and
  Information Science, vol.~705, pp. 133--151. Springer, Cham (2017)

\bibitem{Siwek_2018_FPGA_Abbrev}
Siwek, C., Kowalski, J., Sironi, C.F., Winands, M.H.M.: Implementing
  propositional networks on {FPGA}. AI 2018: Adv. Artificial Intell.
  \textbf{11320},  133--145 (2018)

\bibitem{Soemers_2019_Biasing_Abbrev}
Soemers, D.J.N.J., Piette, {\'E}., Browne, C.: Biasing {MCTS} with features for
  general games. In: Proc. 2019 IEEE Congr. Evol. Computation. pp. 442--449.
  IEEE (2019)

\bibitem{Swiechowski_2015_RecentAdvancesGGP}
{\'S}wiechowski, M., Park, H., Ma{\'n}dziuk, J., Kim, K.J.: Recent advances in
  general game playing. The Scientific World Journal  (2015)

\bibitem{Tak_2012_NGrams_Abbrev}
Tak, M.J.W., Winands, M.H.M., Bj{\"o}rnsson, Y.: N-grams and the
  last-good-reply policy applied in general game playing. IEEE Trans. Comput.
  Intell. AI Games  \textbf{4}(2),  73--83 (2012)

\bibitem{Teytaud_2010_Decisive}
Teytaud, F., Teytaud, O.: {On the Huge Benefit of Decisive Moves in Monte-Carlo
  Tree Search Algorithms}. In: Proceedings of the IEEE Symposium on
  Computational Intelligence and Games. pp. 359--364. Dublin (2010)

\end{thebibliography}
%




\end{document}